\pdfoutput=1

\documentclass[11pt]{article}

\usepackage[]{EACL2023}

\usepackage{times}
\usepackage{latexsym}

\usepackage[utf8]{inputenc} %
\usepackage[T1]{fontenc}    %
\usepackage{hyperref}       %
\usepackage{url}            %
\usepackage{booktabs}       %
\usepackage{amsfonts}       %
\usepackage{nicefrac}       %
\usepackage{microtype}      %
\usepackage{xcolor}         %

\usepackage{graphicx}
\usepackage{makecell}
\usepackage{amsmath}
\usepackage{enumitem}
\usepackage{arydshln}

\graphicspath{ {./fig/} }

\DeclareMathOperator*{\argmax}{\arg\!\max}

\newcommand{\overbar}[1]{\mkern 1.5mu\overline{\mkern-1.5mu#1\mkern-1.5mu}\mkern 1.5mu}

\newcommand{\Lagr}{\mathcal{L}}

\title{Learning to Retrieve Engaging Follow-Up Queries}

\author{Christopher Richardson $^\ddagger$\thanks{\quad Work done during internship.}  \quad 
Sudipta Kar $^\dagger$ \quad 
Anjishnu Kumar $^\dagger$ \quad 
Anand Ramachandran $^\dagger$ \\ 
{\bf Omar Zia Khan} $^\dagger$ \quad 
{\bf Zeynab Raeesy} $^\dagger$ \quad 
{\bf Abhinav Sethy} $^\dagger$\\
$^\ddagger$ Georgia Institute of Technology\\      
$^\dagger$ Amazon Alexa AI\\
\texttt{crichardson332@gmail.com}\\ \{\texttt{sudipkar,anjikum,anramac,ozkhan,raeesyzr,sethya}\}\texttt{@amazon.com}}

\begin{document}
\maketitle

\begin{abstract}
Open domain conversational agents can answer a broad range of targeted  queries. However, the sequential nature of interaction with these systems makes knowledge exploration a lengthy task which burdens the user with asking a chain of well phrased questions. In this paper, we present a retrieval based system and associated  dataset for predicting the next questions that the user might have. Such a system can proactively assist users in knowledge exploration leading to a more engaging dialog. The retrieval system is trained on a dataset called the Follow-up Query Bank (FQ-Bank). FQ-Bank contains $\approx$14K multi-turn information-seeking conversations with a valid follow-up question and a set of invalid candidates. The invalid candidates are generated to simulate various  syntactic and semantic confounders such as paraphrases, partial entity match, irrelevant entity, and ASR errors.
We use confounder specific techniques to simulate these negative examples on the OR-QuAC dataset. 
Then, we train ranking models on FQ-Bank and present results comparing supervised and unsupervised approaches.
The results suggest that we can retrieve the valid follow-ups by ranking them in higher positions compared to confounders, but further knowledge grounding can improve ranking performance.
FQ-Bank is publicly available at \url{https://github.com/amazon-science/fq-bank}.
\end{abstract}

\section{Introduction}
State of the art open domain conversational voice assistants can help users accomplish a wide range of tasks, including: factoid question answering, playing music, adding items to personal lists, controlling smart home appliances, and booking transportation. However, the linear nature of dialog with existing voice assistant technology makes it challenging for users to discover and fully utilize the full range of these capabilities. In addition, successful utilization often requires exact formulation of the request, which further hinders the experience. One recent approach to addressing these issues in the voice assistant domain involves predicting relevant follow-up queries in order to assist the user with accomplishing their latent goals.\footnote{ \url{https://www.amazon.science/blog/alexa-gets-better-at-predicting-customers-goals}}

\begin{figure}
    \centering
    \centering
    \def\svgwidth{\columnwidth}
    \includegraphics[width=\columnwidth]{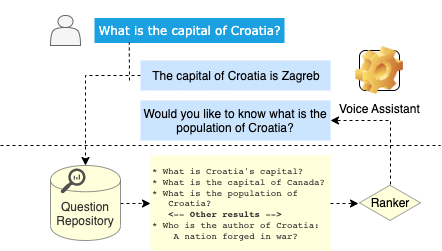}
    \caption{An overview of the follow-up question (FQ) retrieval system.}
    \label{fig:main_diagram}
\end{figure}
\begin{table}[t]
    \centering
    \resizebox{\columnwidth}{!}{
    \begin{tabular}{l}
    \toprule

        \textbf{Dialog History}\\ 
        \makecell[l]{Where was Kurt Gödel born?} \\ \makecell[r]{\textit{Brunn, Austria-Hungary}} \\ \makecell[l]{When was Kurt Gödel born?} \\ \makecell[r]{\textit{April 28, 1906}} \\ \makecell[l]{What was Kurt Gödel’s home life like?} \\ \makecell[r]{\textit{ethnic German family}}\\
        
        \makecell[l]{Where did Kurt Gödel go to school? \\ 
        \makecell[r]{\textit{Godel attended the Evangelische Volksschule in Brunn}}} \\ \midrule
        
        \textbf{Valid Follow up} \\
        What were Kurt Gödel's interests? \\ \midrule

        \textbf{Generated Negative Examples with Types}\\
         Where was Kurt Gödel born?\\
         \makecell[r]{\textit{Duplicate of dialog history}}\\
         
        Which school did Kurt Gödel attend? \\
        \makecell[r]{\textit{Paraphrase}}\\
        
        Where was Cristiano Ronaldo born?\\
        \makecell[r]{\textit{Irrelevant Entity}}\\
 
        Where did Curt Gödel go to school? \\
        \makecell[r]{\textit{ASR Error}}\\
        
        When did Cristiano Ronaldo join Juventus? \\
        \makecell[r]{\textit{Random Question}}\\
        
        When did Kurt Gödel join Juventus? \\
        \makecell[r]{\textit{Irrelevant context}} \\
        \hline

    \end{tabular}}
    \caption{An example showing the dialog history, current turn, valid utterance, and a set of negative utterance candidates from the generated dataset.}
    \label{tab:dataset_example}
\end{table}

Relevant follow-up queries (FQs) for typical voice assistant scenarios can range from specific command and control tasks such as ``\textit{What is the temperature in New York}'' followed by ``\textit{What is the chance of rain in New York?}'', to more open-ended knowledge exploration, e.g. ``\textit{What is the capital of Croatia}" followed by \textit{``What is the population of Croatia?"}. Once a valid FQ has been identified, the system can proactively recommend it to reduce user's cognitive load, e.g. ``\textit{Would you like to know what the population of Croatia is?}''. This exchange is illustrated in Figure \ref{fig:main_diagram}. The user can then be engaged in follow-up dialogue without the need to ask redundant questions.

Follow-up queries can be identified by retrieving and ranking candidates from a question repository. In this approach, we are given dialog of one or more turns between a user and a voice assistant. The system first uses a search engine to retrieve a set of relevant questions by searching against a question repository comprising of historical queries and questions generated from a knowledge base. To create questions from a knowledge base, we can use   templates or use a few-shot Natural Language Generation (NLG) model such as T5 \cite{2020t5}. For example, we can take the tuple {\texttt{\{entity, place of birth\}}} and construct a question template like ``\textit{what is the birthplace of \{entity\}}".

Through a preliminary study of this retrieval approach, we found a basic lexical similarity-based search engine to be ineffective and often returns invalid follow-up queries. Often these top search results included paraphrases of the original query (\textit{When was Cristiano Ronaldo born} $\rightarrow$ \textit{What year was Cristiano Ronaldo born}), as well as similar questions for unrelated entities (\textit{When was Cristiano Ronaldo born} $\rightarrow$ \textit{When was Christian Bale born}). Therefore, an additional ranking module is needed in order to re-rank the search results based on their quality as follow-up queries. To the best of our knowledge, there exists no dataset focused on information-seeking follow-up queries, given a dialog context and a set of valid and invalid follow-up candidates. This problem differs from traditional recommendation systems in that 1) a voice assistant can only recommend one follow-up at a time, and 2) the follow-up query must be highly precise, contextually relevant, and coherent to ensure a positive user experience. This technique can be extended beyond the domain of virtual assistants, for example to chatbots, search engines, and any other smart interaction scenario where contextual coherence and precision is necessary.
Therefore, in this paper, we created the FQ-Bank dataset addressing this problem and explored different modeling techniques to develop a ranking model to retrieve relevant follow-up queries (FQ).
The main contributions of this paper can be summarized as follows:
\begin{enumerate}
    \item For the scenario of a retrieval-based follow-up question selection, we identify a typology of confounders based on preliminary results from a search engine.
    
    \item We propose techniques to synthetically generate confounders according to this typology, based on the publicly available conversation dataset OR-QuAC \cite{qu2020open}, and created the Follow-up Query Bank (FQ-Bank) dataset. FQ-Bank is publicly available and can be used to develop and test machine learning systems for identifying contextually relevant and meaningful follow-up queries from search results. Additionally, the confounder creation techniques can be applied in data augmentation for similar problems. Table \ref{tab:dataset_example} shows an example from the generated dataset.

    \item We adopt a pre-trained language model based approach to develop a benchmark model for ranking a set of candidate follow-up queries for a given factoid utterance and dialog history. We explore the effectiveness of this technique and identify gaps and future directions.
\end{enumerate}
\section{Related Works}
Previous studies on proactivity in conversational AI mostly focuses on response generation.
Follow-up question identification and generation approaches have been explored from different perspectives. For example, \newcite{kundu-etal-2020-learning} explored the task of identifying if the latest user utterance is a follow-up of the previous questions or it has a different new context. This is helpful for understanding the question context properly and give the correct response. They also derived a new dataset called LIF from the QuAC dataset \cite{choi2018quac}, where each data point contains a conversation history, a new utterance, a passage used to answer the previous questions, one valid follow-up, and one or two invalid follow-ups. However, this dataset is focused on passage-based question answering, and the confounder typology does not address issues found in the search engine based  FQ retrieval scenario (e.g. paraphrases, irrelevant entity substitution, etc).

Other works have focused on generating follow-up queries for extracting information from users. For example, \newcite{ge2022should} proposed a knowledge-driven system for generating follow-up queries, but it targeted the generation of follow-up survey questions to extract information from humans. \newcite{Su2018FollowupQG} and \newcite{b-etal-2020-automatic} explored systems for asking follow-up queries to interview candidates to extract more relevant information.

Our proposed method is focused not on information extraction from users, but rather providing highly relevant additional information to the user.

\section{Follow-up Query Bank}\label{sec:data}
For our initial study on identifying FQs, we created a search index of information-seeking questions regarding public facts, spoken by users of a commercial voice assistant. We then queried the search engine with different types of information-seeking questions and analyzed the top negative (i.e., not a suitable follow-up) search results and categorized them into a typology of confounders. Using this typology, we set out the task of simulating a similar scenario on a public conversation dataset and did not use any voice assistant data anymore. We selected OR-QuAC as the seed dataset as it provides multi-turn information-seeking dialogs on a particular topic. 
In the rest of this section, we will provide a brief overview of the confounders and OR-QuAC dataset, followed by an overview of the simulation methodology of the confounders using OR-QuAC.

\subsection{Confounder Identification}\label{sec:confounders}
We created a search index with an open-source search engine for a set of de-identified user interactions with a commercial voice assistant during a period of time.
Then, we carefully selected a set of questions that are different from each other in aspects such as the intent, entity in context, entity's gender and topic domain.
We searched the index against each of these questions and inspected the relevance of the top 20 search results as a follow-up question.
We found that most of ($\approx95\%$) the top search results are not suitable candidates for a follow-up.
We analyzed the top irrelevant results and categorized them as the following confounders that should rank low in their relevance as follow-ups.

\begin{itemize}
    \item \textbf{Paraphrase} We observed a large segment of irrelevant candidates that are semantic equivalents of the query question. This happens because people can ask the same question in different ways. For example, ``\textit{How old is Joe Biden}" and ``\textit{Joe Biden age}" are lexically different but semantically equivalent.
    
    \item \textbf{Irrelevant Entities} 
    Often, the top search results are about entities different from the query question, but the questions have a similar carrier phrase.
    For example, ``\textit{What team does Ronaldo play for}" retrieves questions like ``\textit{What team does Tom Brady play for}". It is true that some user may find such questions as relevant follow-ups, but this is highly subjective. Tom Brady will be completely irrelevant to a user who does not follow the National Football League (NFL). As a result, we considered that such scenarios are irrelevant for now.
    
    \item \textbf{Partial Entity Match} This is a variation of the previous confounder. Here, not only is the carrier phrase similar, but also, the entities share one or more tokens. For example, ``\textit{How old is the University of Washington}" can retrieve questions like ``\textit{How old is the University of Houston}". Here we observe partial entity match ``\textit{university of}" in addition to the identical carrier phrase ``\textit{how old is the}". 
    
    \item \textbf{Irrelevant Context} 
    Some top search results share the correct entity with the query question, but the retrieval can be a non-sequitur.
    For example, ``\textit{What is the capital of France}" can retrieve questions like ``\textit{Where is France}".
    Even though contextually it is a relevant questions, asking back ``\textit{Would you like to know where is France}'' does not make a good experience for the user as the chances are high that the user already have an idea on the geographical position of France.
\end{itemize}

\begin{table}[]
\centering
\resizebox{\columnwidth}{!}{
\begin{tabular}{lrrr}
\toprule
 & \textbf{Train} & \textbf{Validation} & \textbf{Test} \\ \midrule
\textbf{Dialog} & 13,480 & 1,445 & 2,132 \\ 
\textbf{Turns} (utterance, response pair) & 52,712 & 5,534 & 8,195 \\ 
\textbf{Turns per dialog} & 3.91 & 3.86 & 3.84 \\ 
\textbf{Tokens per utterance} & 10.39 & 10.15 & 10.12 \\ 
\textbf{Tokens per response} & 16.76 & 17.00 & 16.90 \\ \midrule

\multicolumn{4}{c}{\textbf{Confounders}}\\ \midrule
\textbf{Paraphrase} & 30,707 & 3,033 & 4,676 \\ 
\textit{- per dialog} & 2.28 & 2.11 & 2.19
\vspace{4pt}\\

\textbf{Irrelevant entity} & 91,490 & 9,660 & 13,736 \\
\textit{- per dialog} & 6.79 & 6.74 & 6.44
\vspace{4pt}\\

\textbf{Irrelevant context} & 51,342 & 5,479 & 8,153 \\
\textit{- per dialog} & 3.81& 3.82 & 3.82
\vspace{4pt}\\

\textbf{ASR Error} & 85,136  & 8,906 & 11,007 \\
\textit{- per dialog} & 6.32 & 6.21 & 5.16 
\vspace{4pt}\\

\textbf{Random utterance} & 40,440 & 4,302 & 6,396 \\
\textit{- per dialog} & 3 & 3 & 3 
\vspace{4pt}\\

\textbf{Duplication of dialog history} & 52,712 & 5,534 & 8,195\\
\textit{- per dialog} & 3.91 & 3.86 & 3.84 
\vspace{4pt}\\
\midrule

\textbf{Total} & 3,51,827 & 36,914 & 52,163\\
\textit{- per dialog} & 26.10 & 25.74 & 24.47 \\
\bottomrule

\end{tabular}%
}
\caption{Statistics of the created dataset with each category of the negative examples.}
\label{tab:data_stat}
\end{table}
Additionally, we listed the following confounders that were not seen in our limited data analysis but can appear in a larger system:

\begin{itemize}
    \item \textbf{Automatic Speech Recognition (ASR) Error} ASR failures can sometimes replace an entity with a similar sounding word. For example, \textit{Kurt} can be replaced with \textit{Curt} in "\textit{Where did Kurt Gödel go to school}". High lexical overlap can rank such irrelevant entities highly.
    
    \item \textbf{Duplication of Dialog History} Sometimes the information provided by a candidate follow-up question can already be present in a multi-turn dialog history. In such a case, it is important to identify and get rid of those questions by modeling the dialog history.

\end{itemize}

After identifying these confounder categories, we selected OR-QuAC as the starting dataset, and used different techniques to generate these confounders and simulate the retrieval scenario for the follow-up selection system.

\subsection{OR-QuAC Dataset}
Open-Retrieval Conversational Question Answering (OR-QuAC) consists of $\approx$6K multi-turn information-seeking dialogues between two humans, one posing as student (asks knowledge-seeking questions) and the other as teacher (answers the questions using Wikipedia as the knowledge source).
It draws from the popular QA dataset QuAC \cite{choi2018quac} as well as CANARD \cite{ghoneim2019canard}, which provides context-independent rewrites of initial questions written by human annotators. 

This dataset is well-suited for our purposes as: i) the conversations aim at exploring knowledge about entities or topics, ii) multi-turn conversations enable us simulating a dialog history (one or more question-answering turns between two people), iii) query rewrites are helpful to get rid of anaphoric references which can make candidate questions ambiguous about entities (e.g., ``\textit{How many kids Kamala Harris has}" removes the ambiguity from ``\textit{How many kids she has}").

For each information-seeking question in the OR-QuAC dataset, we chose the rewritten version as the current question, the previous turns as the dialog history, and the immediate next turn as the valid follow-up question. 
Then, we used different techniques to generate the confounders that we will explain in the next section.

\subsection{Data Sample Generation}
For a conversation in the OR-QuAC dataset of $T$ turns (question$-$answer pairs), we have sampled $T{-1}$ data points $\{x, y\}$.
Each generated data sample contains a dialogue context, $x$, of length $\Lagr$ ($1{\leq} \Lagr {\leq{T{-}1}}$). $x$ contains a dialog history $\{(q_1, a_1), ....,$ $(q_{\Lagr-1}, a_{\Lagr-1})\}$ of length $\Lagr-1$ (i.e., $\Lagr{-1}$ question ($q$)$-$answer ($a$) pairs), and a current question ($q_\Lagr$) and the answer ($a_\Lagr$). Hence, \mbox{$x = \{(q_1, a_1), ...., (q_{\Lagr-1}, a_{\Lagr-1}), (q_{\Lagr}, a_{\Lagr})\}$}.

Each data sample also contains a set of positive and negative follow-up queries, $y = \{y^+\} \cup y^-$. $y^+$ is a single positive follow-up question, and $y^-$ is a set of negative follow-ups ($y^- = \{y_1^-, y_2^-, ... \}$) that we have created based on the identified confounders.

\textbf{Valid Examples}
Given a dialog history $x_{1:T-1}$ of length $T{-1}$ and a current turn $x_T$, we consider the consecutive question ($x_{T+1}$) in the OR-QuAC dataset as a positive follow-up question.

\textbf{Adversarial Examples} We used the following methods to populate the candidate question space for a turn with negative examples based on the confounders we have listed in Section \ref{sec:confounders}:
\begin{itemize}
    \item \textbf{Paraphrase:} We used a pre-trained BART model \cite{lewis2019bart} that was fine-tuned on several paraphrase datasets\footnote{\url{https://huggingface.co/eugenesiow/bart-paraphrase}}. For the last user turn in a dialog history, we used this model generated paraphrase as a confounder.
    
     \item \textbf{Irrelevant Entities and Partial Entity Match:} We first used the SpaCy\footnote{\url{http://spacy.io}} library to identify the named entities in the current question in a turn. Then we generate a negative example by replacing the entity with an entity of a similar type from a catalog generated from WikiData. For entities with multiple word tokens, we replace a token (e.g., first name or last name) with a random first name or last name token. For a dialog, we created multiple such examples.
    
    \item \textbf{Irrelevant Context:} We randomly sampled one question from the rest of the dataset that has a similar entity type and replace that with the entity in the context of a current question. That means, for an entity we swap the original question with a %
    random one. 
    
    \item \textbf{Random question:} We added three %
    random questions from the dataset as a negative examples for a dialog.
    
    \item \textbf{ASR Error:} For an entity in a question, we generated a similar sounding entity using the Datamuse API \footnote{\url{https://www.datamuse.com/api}} and replaced the original entity with the generated homophone. For entities with multiple word tokens, we created multiple examples like this by replacing one token with a homophone at a time. 
    
    \item \textbf{Duplication of Dialog History:} We added a question from the dialog history in the candidate set.
\end{itemize}

We maintained the standard training, validation, and test splits from OR-QuAC while generating the dataset. Table \ref{tab:dataset_example} shows an example of generated data and Table \ref{tab:data_stat} shows statistics of the dataset.
As we maintained the original data split, distribution is similar across all the splits. 
For each dialog, there are $\approx$25 negative examples with one positive example. 
That means, a model needs to learn contextual relevancy for being able to identify the correct follow-up.

\section{Learning to Identify Relevant Follow-up}\label{sec:methodology}

\noindent\textbf{Task Formulation:} 
Given a dialog $x=\{ (q_1, a_1), ...., $ $(q_{\Lagr-1}, a_{\Lagr-1}), (q_{\Lagr}, a_{\Lagr})\}$ of $\Lagr$ turns and a randomly organized set of $n$ candidate follow-up queries $y = \{y^+\} \cup \{y^{-}_1, y^{-}_2, ..., y^{-}_{n-1}\}$, the task is to model $P(i|x), i \in y$, such that $\argmax_i P(i|x) = y^+.$\\

Here, $q$ is a question, $a$ is an answer, $y^+$ is a positive follow-up example, and $y^-$ is a set of negative examples.
In order to develop a follow-up question candidate ranker, we experiment with different unsupervised and supervised approaches as described below.

\subsection{Unsupervised}
We experimented with Glove \cite{pennington2014glove} word embeddings, pre-trained SentenceBERT \cite{reimers2019sentence} model in the unsupervised direction.
For a given dialog $x$ and candidate utterance set $y = \{y^+, y^-\}$, we use the Glove or SentenceBERT to generate a high-level vector representation $\overbar{x}$ from the dialog and do the same for each of the candidate utterances $y_i \in y$.
With Glove, we compute the mean of the $300d$ embedding vectors for all the word tokens in a dialog $x$ and represent the out-of-vocabulary (OOV) words with zero vectors.
The vocabulary coverage of Glove is $\approx$99\% for the dataset.
For SentenceBERT, we feed the entire input texts (concatenation of multiple turns in $x$) for $x$ and $y_i \in y$ to generate $\overbar{x}$ and $\overbar{y_i}$, respectively.
Then, we compute the cosine similarity $\alpha = \cos(\overbar{x}, \overbar{y_i})$ between $\overbar{x}$ and $y_i \in y$ and rearrange $y$ in descending order based on $\alpha$.

\subsection{Supervised}
For the supervised experiments, we fine-tune a pre-trained language model by translating the problem as a binary classification task.
In other words, for a given dialog \mbox{$x = \{ (q_1, a_1), ...., (q_{\Lagr-1}, a_{\Lagr-1}), (q_{T}, a_{T})\}$} and a candidate set $y = \{y^+\} \cup y^-$, we train a model $\theta$ to predict $\hat{y} = P(i \mid x), i \in y$, where $\hat{y} \rightarrow \mathbb{R}: [0,1]$.

We format the input by concatenating the dialog history turns and a candidate utterance with a \texttt{[SEP]} token and a single output node outputs a continuous value between 0 and 1.
As the starter pre-trained language model we experiment with BERT \cite{devlin-etal-2019-bert} and RoBERTa \cite{liu2020roberta}.
We use the bert-base-cased\footnote{\url{https://huggingface.co/bert-base-cased}} and roberta-base\footnote{\url{https://huggingface.co/roberta-base}} variations of these models.
We fine-tune the models for 20 epochs with an early stopping patience of three epochs with a learning rate of $2e^{-5}$ and batch size of 64.
We use cross-entropy loss to optimize the model with AdamW optimizer.
During inference, we use the model predicted score to rearrange the candidate set for a dialog in descending order.

\section{Experiments and Results}
\paragraph{Evaluation Metric:} As the task is to rank the valid follow-up question higher than a set of invalid confounders, we evaluate the performance using Mean Reciprocal Rank (MRR), given as:

\begin{align}
    MRR = \frac{1}{|Q|} \sum_{i=1}^{|Q|} \frac{1}{\text{rank}_i}
\end{align}
where $\text{rank}_i$ is the rank position of the valid candidate for the $i$th datapoint.

Additionally, we compute Hit Ratio@1 and Hit Ratio@3 for the top performing methods to analyze the percentage of samples for which the ranking method ranked the correct candidate as the first item and within the first three items.

\paragraph{Quantitative Results:}
\begin{table}[t]
\centering
\resizebox{0.95\columnwidth}{!}{%
\begin{tabular}{lrr}
\hline

 &  \textbf{Validation} & \textbf{Test} \\ \hline
 
\multicolumn{3}{l}{\textbf{Unsupervised}} \\ 
Glove & 0.142 & 0.141 \\ 
SentenceBERT  & 0.133 & 0.141 \\ \hline

\multicolumn{3}{l}{\textbf{Supervised}} \\ 
BERT &  0.842 & 0.805 \\ 

RoBERTa & 0.838 & 0.808 \\  \hdashline
\multicolumn{3}{l}{Hit Ratio@1/ Hit Ratio@3} \\ 
BERT & 72.0/ 89.3 & 68.5/ 88.1\\ 
RoBERTa & 71.7/ 88.7 & 68.2/ 89.5\\
\hline
\end{tabular}
}
\caption{Ranking performance in MRR for different unsupervised and supervised methods. The last two rows show the Hit Ratio at the first and third position for BERT and RoBERTa.}
\label{tab:main-results}
\end{table}
In Table~\ref{tab:main-results} we report   results comparing the methods proposed in Section~\ref{sec:methodology} on the adversarial dataset described in Section~\ref{sec:data}.  
 The unsupervised methods performed poorly for the ranking task resulting into MRR scores of 0.141 for both Glove and SentenceBERT based embeddings, and the trend is similar for all the data splits.
This is not surprising as cosine similarity is expected to be high for paraphrases. As discussed in section~\ref{sec:confounders}, paraphrases are not good candidates for FQs as they don't provide any value to the user.

\begin{table}[t]
    \centering
    \resizebox{0.9\columnwidth}{!}{
        \begin{tabular}{p{0.9\columnwidth}}
        \toprule
        \textbf{Dialog context}
        
        \begin{itemize}[noitemsep]
            \item Where was Michael Bennett born?
            \item Who are Michael Bennett's parents?
            \item When did Michael Bennett's career begin?
            \item What show did Michael Bennett begin his career?
        \end{itemize}

            \textbf{Irrelevant Context Candidate}\\
            \textit{When did Michael Bennett move to Alaska?} (Model score: 0.3)\\ 
            
            \textbf{Valid Candidate}\\
            \textit{What was Michael Bennett's role in the "Here's Love" and "Bajour"?} (Model score: 0.2)\\ 
        
        \midrule
        \textbf{Dialog context}
        \begin{itemize}[noitemsep]
            \item What happened to Sachin Tendulkar during the tour of Australia? 
            \item How did Sachin Tendulkar do in the 2003 Tour of Australia?
            \item How many games did Sachin Tendulkar win during 2003?
            \item Did Sachin Tendulkar win any awards?
        \end{itemize}

            \textbf{Irrelevant Context Candidate}\\
            \textit{How many hits did Sachin Tendulkar have?} (Model score: 0.21)\\              \textbf{Valid Candidate}\\
            \textit{Was there any controversies for Sachin Tendulkar?} (Model score: 0.35)\\ 
        \bottomrule
        
        \end{tabular}
    }
    \caption{Examples where the model predicted scores do not match with the category of the follow-ups.}
    \label{tab:error_example}
\end{table}

We observe a large improvement when we fine-tune pre-trained language models like BERT and RoBERTa to simply classify each candidates as relevant or irrelevant.
The MRR is $\approx0.8$ when we treat the models' confidence score for relevancy as the basis for ranking the candidates.

\begin{figure}
    \centering
    \includegraphics[width=0.9\linewidth]{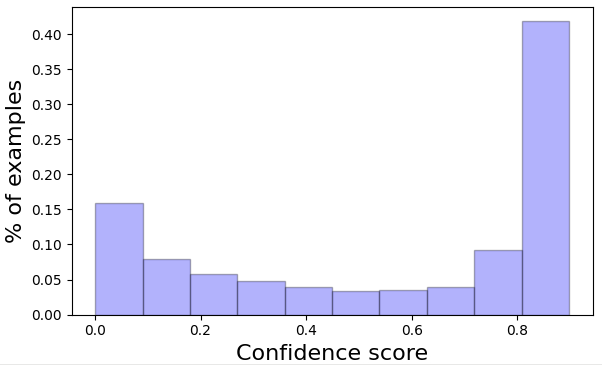}
    \caption{Histograms of model predictions for valid follow-up queries.}
    \label{fig:hist_valid}
\end{figure}

The Hit Ratio@1 and Hit Ratio@3 metrics show that both BERT and RoBERTa ranked the correct FQ at the first rank for $\approx$68\% cases. Both the models ranked the correct FQ within the first three items for $\approx$88-89\% cases. This shows promise in using such methods to retrieve a relevant follow-up question.

\paragraph{Error Analysis} Analysing the fine-tuned models' predicted scores for different confounder types, we have found that the models can identify most of the confounder types easily.
For example, the model predicted score is below 0.1 for $\approx99\%$ candidates from duplication of dialog history, ASR errors, and random utterance confounder.
However, the models often predicted higher scores for the candidates from the irrelevant context category (score is $<0.1$ for 83\% candidates).
For 8\% of these candidates, the score is higher than 0.4, which is not the case with other categories of confounders.

Inspecting some examples like the ones presented in Table \ref{tab:error_example}, we found that without having factual information about the entities or topics, such irrelevant contexts are often difficult to identify for humans as well.
They can look linguistically plausible but have factual errors.
For example, the question ``\textit{How many hits did Sachin Tendulkar have}'' sounds plausible to some humans, but is actually invalid. Tendulkar is a cricket player, and `hits' is not a statistic in cricket. However, it is a real statistic in baseball, so specific domain knowledge is needed to rule this out as a valid FQ. This example illustrates how integrating information about entities from knowledge bases can be helpful for the system, ans this method can be explored in the future. Although scores like 0.2 do not look very high in general for a scale of 0 to 1, Figure \ref{fig:hist_valid} shows that the model assigns such scores to a large portion of valid follow-up queries.

Observing the model's overall performance (MRR of $\approx$0.8) in ranking the valid candidates at a better position than the invalid ones shows promise in using such a system can be a good starting point for developing a follow-up question retrieval system. A large advantage of the proposed adversarial example generation methods and the proposed dataset is that these can help to bypass the need for exhaustive data annotation need for developing a follow-up generation system. Additionally, the trained model using this dataset can be further fine-tuned by annotating a small number of case specific examples, which would help to improve the model accuracy and adapt in different use cases, as well as reach a higher accuracy in identifying suitable follow-up queries.

\section{Conclusions}
In this paper, we sought to address the problem of identifying valid and engaging follow-up queries for a user interacting with a conversational assistant. We experimented with a retrieval and ranking based framework to achieve this using a search engine and a database of past queries. In doing so, we identified a typology of confounders returned by the search. In order to train a ranking model to identify valid follow-up queries, we synthetically generated confounders based on a publicly available conversation dataset. We showed that our approach of ranking retrieved candidates based on their validity as follow-up queries achieved reasonable performance, but also that integrating external knowledge on entities or topics could improve follow-up selection. We have made the dataset publicly available to enable further research in this direction.

\section{Limitations}
The first limitation of this work is that we are attempting to mimic conversational interactions with publicly available human-annotated data based on Wikipedia. Thus in some cases the generated dataset can contain dialogues unrealistic to the voice assistant scenario. Additionally, despite our typology of confounders being based on results from a search-based approach using real data, there are inevitably additional types of potential confounders not fully covered by our approach.

Second, we only focused on contextual relevance and coherence through the lens of language. But, in practice, there are external factors like user preference, time of the day, repetition in a longer period (e.g., a user may have asked the question in the follow-up a couple of days ago and it does not make any sense to ask the same question as a follow-up). More comprehensive methods would be needed to address these concerns.

Finally, this dataset is limited to knowledge-seeking queries. Other types of valid follow-up actions (e.g. setting a timer, booking a ride) are not included in this dataset.

\bibliography{anthology,custom}
\bibliographystyle{acl_natbib}

\end{document}